\documentclass{article}

\usepackage[final]{corl_2026} 

\usepackage{hyperref}       
\usepackage{url}            
\usepackage{booktabs}       
\usepackage{amsfonts}       
\usepackage{nicefrac}       
\usepackage{microtype}      
\usepackage{graphicx}
\usepackage{subcaption}
\usepackage{hyperref}
\usepackage{amsmath}
\usepackage{amssymb}
\usepackage{mathtools}
\usepackage{amsthm}
\usepackage{multirow}
\usepackage{wrapfig}
\usepackage{bm}
\usepackage{algorithm}
\usepackage{algorithmic}
\usepackage{needspace}     

\newcommand{\sO}{\mathcal{O}}
\newcommand{\sS}{\mathcal{S}}
\newcommand{\sA}{\mathcal{A}}

\newcommand{\sD}{\mathcal{D}}

\newcommand{\sL}{\mathcal{L}}

\newcommand{\sX}{\mathcal{X}}
\newcommand{\E}{\mathbb{E}}

\title{RoboFQE: Rollout-Free Evaluation of Visual Robot Policies from Offline Trajectories}

\author{
  Yunfu Deng \quad Sangwoo Shin \quad Brahma S.~Pavse \quad Josiah P.~Hanna\\
  Department of Computer Sciences\\
  University of Wisconsin--Madison\\
  \texttt{yunfu.deng@wisc.edu} \quad \texttt{sshin237@cs.wisc.edu}\\
  \texttt{pavse@wisc.edu} \quad \texttt{jphanna@cs.wisc.edu}
}

\begin{document}
\maketitle
{\let\thefootnote\relax\footnotetext{Project page: \href{https://yfdeng.com/robofqe/}{\texttt{yfdeng.com/robofqe}}}}

\begin{abstract}
%
    \looseness=-1 Estimating the performance of a robot's control policy before real-world deployment typically relies on rollout-based approaches, which execute the policy in a simulator or in a learned world model.
    Both approaches are costly to construct: simulators require extensive domain expertise, while world models require large-scale training infrastructure and additionally suffer from prediction errors that compound over the evaluation horizon.
    To address these challenges, we leverage off-policy evaluation (OPE), a framework that estimates the return of a target policy from a fixed dataset of offline trajectories. 
    OPE has been studied primarily on low-dimensional, state-based tasks, and its effectiveness for visuomotor robot control remains largely unexplored.
    To make OPE practical in this setting, we introduce RoboFQE, 
    which fits a Q-function to offline trajectories, without ever rolling out the policy. RoboFQE leverages a pretrained vision foundation model to handle high-dimensional inputs and a learned terminal success detector to supply reward signals without labels.
    %
    %
    %
    %
    Through experiments on simulated and real-world manipulation tasks, we demonstrate that RoboFQE produces reliable return estimates, significantly outperforming world-model evaluators by a growing margin as the evaluation horizon increases.
    We further show that the pretrained vision foundation model is critical for RoboFQE on long-horizon tasks, while the learned success detector eliminates the need for ground-truth reward labels with minimal loss in accuracy.
\end{abstract}


\section{Introduction}

Generalist robot policies are advancing rapidly. Recent vision-language-action models, diffusion-based policies, and large-scale robot learning systems increasingly aim to operate across tasks, objects, scenes, and embodiments~\citep{team2024octo,kim2024openvla,black2026pi0visionlanguageactionflowmodel,bjorck2025gr00t,liu2025rdt,shukor2025smolvla,chi2025diffusion}. This progress creates a fundamental but unresolved evaluation problem: how should such policies be evaluated? The most direct approach is repeated real-world testing. A policy can be executed across many tasks, initial states, objects, and environments, with enough repeated trials to obtain statistically meaningful success estimates. However, this gold-standard procedure is expensive and difficult to scale. Each trial may require scene setup, resets, failure recovery, hardware operation, and human supervision. Because these costs multiply with the number of candidate policies and development iterations, statistically meaningful evaluation becomes a bottleneck during policy development.

Recent work has made substantial progress by scaling how evaluation rollouts are obtained. Automated real-world evaluation systems reduce human labor by automating scene resets and success detection~\citep{zhou2025autoeval}, 
and distributed evaluation protocols pool real-robot trials from multiple laboratories into unified policy rankings~\citep{atreya2025roboarena}.
Simulator-based and real-to-sim methods evaluate policies in reconstructed environments, sometimes with statistical calibration to real-world outcomes~\citep{li24simpler,badithela2025reliable}. More recently, action-conditioned video world models have been used to simulate the evaluation environment, where generated policy rollouts can be scored automatically~\citep{quevedo2026worldgym,tseng2025scalable,veorobotics2025,guo2026ctrl}. These approaches share a common structure in which policy performance is estimated by producing new trajectories, whether in the real world, in a simulator, or in a learned model.

\begin{figure*}[t]
    \centering
    \includegraphics[width=\textwidth]{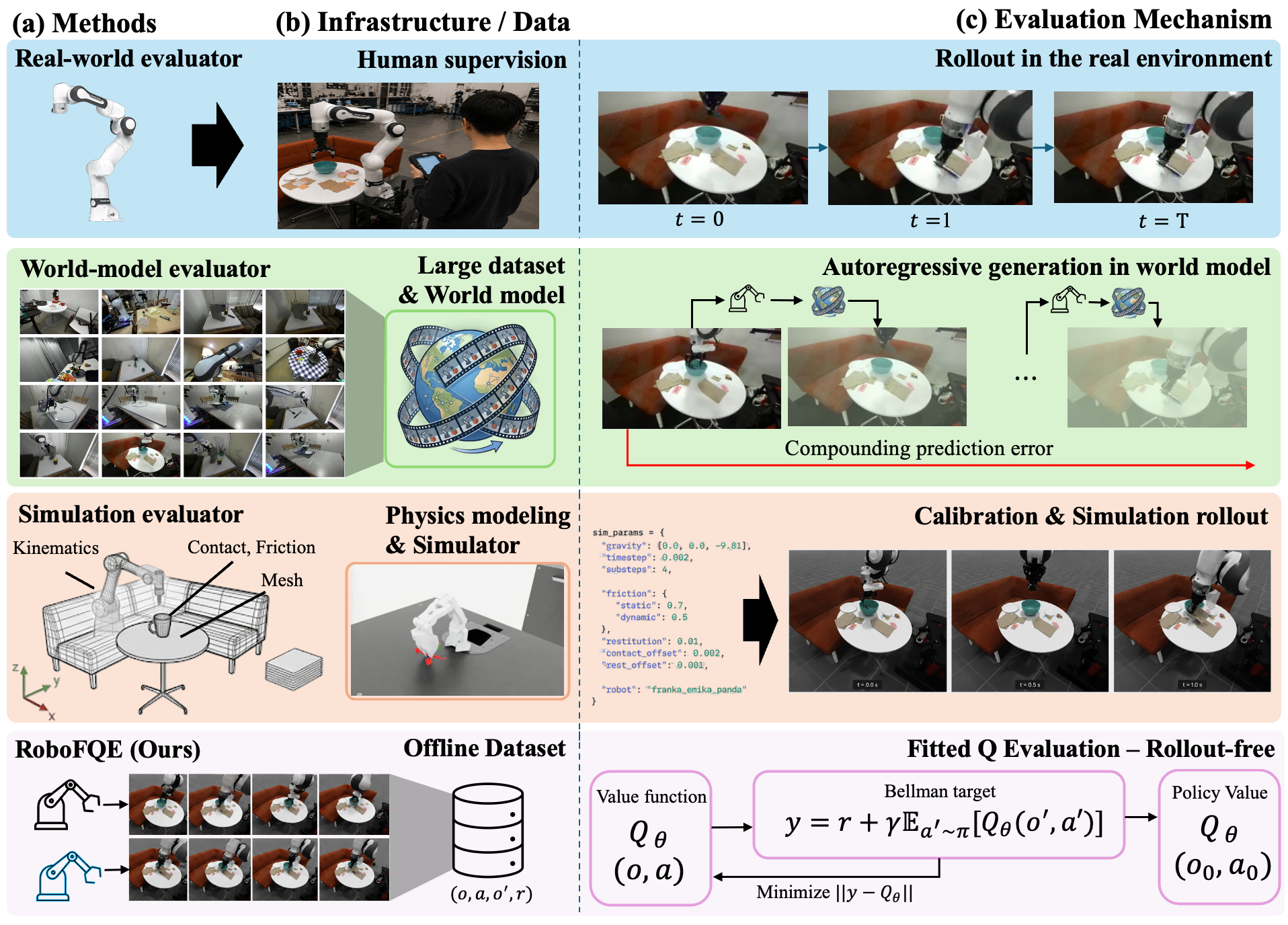}
    \caption{Comparison of robot policy evaluation methods.}
    \label{fig:overview}
\end{figure*}

However, as shown in Figure~\ref{fig:overview}, each of these rollout-based paradigms imposes its own requirements in visual robot manipulation. Real-world rollouts require physical access and repeated interaction with the deployment environment under human supervision. 
Simulator-based evaluation requires environment reconstruction, system identification, or calibration, and remains affected by sim-to-real mismatch. World-model-based evaluation avoids hand-built simulators but relies on autoregressive trajectory generation, and prediction errors can compound over long manipulation sequences~\citep{sutton1996model,xiao2019learning}. 
These costs motivate evaluation methods that avoid rollouts of the evaluated policy entirely.
Value-based off-policy evaluation (OPE) addresses this need by estimating the expected return of a target policy by iteratively fitting a value function to one-step Bellman targets constructed from offline transitions, 
recovering long-horizon returns without any additional environment interactions~\citep{li2019perspective}. 
The offline data OPE requires is abundant in robotics: teleoperation demonstrations, policy rollouts, and open-source datasets accumulate as policies are trained. OPE reuses this data, whereas simulators and world models must be built before a policy can be evaluated.

However, OPE has been studied primarily on low-dimensional, state-based benchmarks~\citep{voloshin2021empirical,fu2021benchmarks}, 
and its effectiveness 
in visual robot manipulation remains an open empirical question.
In this paper, we address this question by introducing RoboFQE, a rollout-free OPE evaluator augmented with pretrained visual representations and learned reward annotation. 
Among different categories of OPE, RoboFQE builds on fitted Q-evaluation (FQE)~\cite{ernst2005tree,pmlr-v97-le19a}, 
a value-based OPE method that learns the target policy's Q-function by regression.
By avoiding both a dynamics model and estimation of the data-collection policy's action distribution, FQE is simpler than other OPE methods and ranks among the most accurate estimators in prior empirical comparisons~\citep{voloshin2021empirical,fu2021benchmarks}. 
To further adapt FQE to the visual robot domain, we introduce two design choices.
First, a pretrained vision foundation model serves as the Q-function's representation backbone, efficiently handling the high-dimensional visual observations that existing OPE benchmarks rarely cover. 
Second, a learned terminal success detector supplies reward annotations for trajectories without ground-truth labels. As offline robotics data often lacks reward signals or outcome labels, conflicting with the standard OPE assumption that rewards are given in the data, we introduce this learned success detector to bridge the gap.

Our experiments on the LIBERO and CALVIN benchmarks~\citep{liu2023libero,mees2022calvin} show that RoboFQE achieves the lowest absolute return error across four manipulation task suites, particularly on long-horizon tasks where rollout-based world-model evaluators degrade.
Specifically, RoboFQE reaches only 5.63\% return error on the longest-horizon suite while the strongest world-model evaluator exhibits 54.4\%, as autoregressive prediction errors compound over horizons. This result suggests that our rollout-free OPE evaluator will become increasingly relevant as foundation policies tackle longer and more complex task sequences.
Beyond simulation, RoboFQE also achieves lower return-estimation error than world-model evaluators on a real robot, even with its success detector applied zero-shot.

We summarize our contributions as follows: \textbf{(i)} To the best of our knowledge, we present the first systematic study of value-based OPE for visual robot manipulation, and introduce RoboFQE as a practical instantiation of FQE for this domain by incorporating a vision foundation model and a learned success detector into FQE. \textbf{(ii)} Through experiments on the LIBERO and CALVIN benchmarks and on real-world manipulation tasks, we show that RoboFQE achieves lower return-estimation error than rollout-based evaluators.

\section{Related Work: Evaluation Paradigms and Data Constraints}
\label{sec:evaluation_paradigms}

\paragraph{Rollout-based robot policy evaluation.}
\label{sec:rollout_based_eval}
While direct  real-world execution remains the gold standard for evaluating robot policies, its cost has motivated alternative paradigms.
One class of alternatives replaces real-world execution with simulator-based proxies. 
Since their evaluation accuracy depends on how closely the simulator matches the real deployment environment,
the central challenge lies in faithfully reconstructing real evaluation scenes within the physics simulator. 
\citet{li24simpler} address this challenge through system identification and texture-baked rendering, while \citet{jain2025polarisscalablerealtosimevaluations} use 3D Gaussian splatting.
To account for the inevitable mismatch between simulator and real-world outcomes, \citet{badithela2025reliable} combine imperfect simulators with limited real-world data through a prediction-powered inference framework~\citep{angelopoulos2023prediction}.

\looseness=-1 More recently, action-conditioned video world models have been proposed as potential evaluation protocols. 
The core idea is to roll out the target policy inside a generative video model rather than the real world or a hand-built physics simulator.
For example, \citet{quevedo2026worldgym} train an autoregressive, diffusion-based video generation model on Open X-Embodiment real robot data and score rollouts with a vision-language model.
Subsequent work scales this approach by leveraging foundation-model pretraining:
\citet{tseng2025scalable} initialize from the Cosmos-Predict model~\citep{ali2025world} and augment with policy rollouts, while
Gemini robotics team~\citep{veorobotics2025} build on the frontier Veo~\citep{veoteam2025veo} model with multi-view conditioning, validated across over 1,600 real-world evaluation episodes. 
Recent designs explicitly target consistency over long rollouts: \citet{guo2026ctrl} use pose-conditioned memory retrieval, and concurrent work proposes multi-view latent world models~\citep{jain2026weaver} and self-consistent video generation~\citep{tseng2026sc3eval}.

Although both paradigms offer promising alternatives to real-world hardware execution, the engineering overhead and sim-to-real mismatch of simulation-based approaches and the compounding errors of world-model evaluators remain key bottlenecks for scalable policy evaluation. Our work instead reuses offline trajectories through value-based off-policy evaluation, avoiding both environment construction and trajectory generation.

\paragraph{Offline policy evaluation.}
\label{sec:value_based_ope}
Off-policy evaluation (OPE) estimates the expected return of a target policy from a fixed dataset of offline trajectories, without new environment interactions.
Existing methods broadly fall into three families.
\textit{Importance-sampling and doubly robust estimators}~\citep{pmlr-v48-jiang16,pmlr-v48-thomasa16} reweight offline trajectories by the target-to-behavior policy ratio, but require access to behavior policy action probabilities, which are rarely logged in robot datasets. 
\textit{Stationary distribution correction methods}~\citep{nachum2019dualdice,yang2020off} avoid explicit behavior-policy access but require estimating density-correction ratios through minimax optimization over the joint state-action space, a step that has not been validated in high-dimensional visual observation spaces. 
\textit{Value-based estimators}~\citep{ernst2005tree, pmlr-v97-le19a} learn the target policy's Q-function directly from offline transitions, requiring neither behavior-policy access nor distribution-ratio estimation. 
The DOPE benchmark~\citep{fu2021benchmarks} finds value-based approaches among the most accurate OPE estimators on continuous control tasks. Based on this observation, we adopt fitted Q-evaluation~\citep{pmlr-v97-le19a}, a representative value-based estimator, as the base method for studying OPE in visual robot policy evaluation.

Despite extensive study on low-dimensional benchmarks, OPE has seen limited adoption as a standard evaluation tool in robotics. 
\citet{DBLP:journals/corr/abs-1906-01624} apply OPE to image-based robotic grasping tasks, observing that importance-sampling and learned-dynamics OPE both struggle in visual settings and proposing a classification-based metric for policy ranking. 
\citet{voloshin2021empirical} extend OPE benchmarking to vision-based domains, but these environments consist of rendered Atari and gridworld tasks with less visual complexity than real manipulation scenes. 
\citet{pavse2023state} and \citet{pavse2025stable} develop OPE-specific representation learning to stabilize value estimation under visual observations, but evaluate primarily in state-based domains. 
Our work bridges these gaps by augmenting classical FQE with modern pre-trained visual representations and VLM-based reward annotation, providing accurate value estimation beyond policy ranking while removing the standard OPE assumption of given rewards.
The reward annotation can draw on learned robotic reward models that judge task success from observations, such as LRM~\citep{wu2026large} or Robometer~\citep{liang2026robometer}.

\section{Rollout-Free Evaluation from Offline Robot Trajectories}
\label{sec:rollout_free_eval}

This section formulates the problem of rollout-free policy evaluation from offline robot trajectories and introduces RoboFQE, the instantiation of FQE that we study in this work.

\subsection{Problem Setup}
\label{sec:problem_setup}
We consider language-conditioned visual robotic manipulation tasks, where the agent receives image-based (or multimodal) observations, outputs continuous actions, and receives a sparse terminal reward indicating task success. The task is modeled as a finite-horizon partially observable Markov decision process (POMDP) with latent state $s_t \in \sS$, observation $o_t \in \sO$, action $a_t \in \sA$, natural-language task description $l \in \sL$, horizon $H$, discount factor $\gamma \in (0,1]$, and terminal success function $R_l:\sS \to \{0,1\}$. The latent state is not directly observed; instead, the agent receives observations generated from the underlying state. Because the state is not observed, an optimal policy in general depends on the full observation--action history $h_t = (o_0, a_0, o_1, \dots, a_{t-1}, o_t)$~\citep{kaelbling1998planning}.

In practice, robot policies condition on a task-specific subset of the available information at each timestep, which may include a single image, a fixed window of recent frames, proprioceptive state, multiple camera views, or multimodal combinations thereof. We denote the policy input at timestep $t$ as $x_t \in \sX$, where $x_t = f(h_t)$ is a deterministic function of the history. A target policy is then written as $\pi: \sX \times \sL \to \Delta(\sA),$ with $
a_t \sim \pi(\cdot \mid x_t, l).$

The reward is sparse and terminal: $R_l(s_{t+1}) = 0$ for all $t < T - 1$, and $R_l(s_T) \in \{0,1\}$ indicates task completion, where $T \leq H$ is the termination time of the episode. 
The evaluation objective is the expected discounted return of $\pi$ under task $l$,\footnote{The objective is defined per task $l$; for a generalist policy operating over a task distribution $p(l)$, it extends directly to $V^\pi = \E_{l \sim p}[V_l^\pi]$.}
\begin{equation}
V_l^\pi \;\triangleq\; \E_{\pi}\!\left[\, \sum_{t=0}^{T-1} \gamma^t\, R_l(s_{t+1}) \,\right] \;=\; \E_{\pi}\!\left[\, \gamma^{T-1}\, R_l(s_T) \,\right].
\label{eq:policy_value}
\end{equation}

Being able to estimate $V_l^\pi$ supports both policy ranking and threshold-based deployment decisions, as it indicates not only which policy is better, but also whether a given policy's expected return exceeds a minimum acceptable level.

\subsection{Off-Policy Evaluation Setting}
\label{sec:ope_setting}

The estimator has access to a fixed offline robot trajectory dataset:
\begin{equation}
\sD = \{\tau_i\}_{i=1}^{N},
\qquad
\tau_i = \big(\, l_i,\; o^{(i)}_0,\, a^{(i)}_0,\, o^{(i)}_1,\, a^{(i)}_1,\, \ldots,\, a^{(i)}_{T_i - 1},\, o^{(i)}_{T_i}\,\big),
\label{eq:dataset}
\end{equation}
where $l_i \in \sL$ is the language description of the task attempted in trajectory $i$, and $T_i$ is the trajectory length. The dataset contains observations and actions, but not latent states. The observations may include RGB images, multiple camera views, proprioception, or gripper states. For each trajectory, the policy input $x^{(i)}_t$ is constructed from the recorded data according to the input function $f$ used by the target policy.

This setting poses three challenges for value estimation:
\textit{(i)} Reward annotations may be absent. While most OPE work assumes that a reward $r_t^{(i)}$ is recorded at every timestep $t = 0,\ldots,T_i-1$ of each trajectory $\tau_i$, real-world robot datasets typically record only observations and actions, with no annotation of task success.
When ground-truth success labels are available, the estimator uses them directly; otherwise, task success must be inferred from the terminal observation and language instruction using a success detector $\hat{R}_l: \sO \times \sL \to \{0,1\}$, and the discrepancy between $\hat{R}_l(o_T,l)$ and $R_l(s_T)$ introduces reward-inference error into the downstream value estimate.
\textit{(ii)} The action distribution of the data-collection process is generally unknown. Real-world robot data are often aggregated across teleoperators, scripted controllers, earlier policy versions, and source datasets, so the probability with which each recorded action was taken is neither stationary nor available in closed form. RoboFQE therefore assumes no access to these probabilities, though it can operate in settings, such as scripted simulator benchmarks, where they are known.
\textit{(iii)} No new evaluation rollouts are available. The target policy may be queried for actions, but it is not executed in the real environment, in a hand-built simulator, or in a learned world model.

\subsection{Value Estimation via Fitted Q-Evaluation}
\label{sec:robofqe}

RoboFQE estimates the discounted expected return of the target policy by fitting an action-value function to one-step Bellman targets constructed from the offline dataset $\sD$. The estimator operates on offline data and target policy action samples, without learning a dynamics model or estimating a behavior-policy density.

First, each trajectory is annotated once with a terminal success label $\hat{r}_i \in \{0,1\}$, which is shared by all target policies and given by the success detector $\hat{R}_{l_i}(o^{(i)}_{T_i}, l_i)$ when ground-truth labels are unavailable.
Annotating $\sD$ with terminal labels and constructing policy inputs yields the reward-augmented transition dataset:
\begin{equation}
\sD' \;=\; \big\{\, (x^{(i)}_t,\; a^{(i)}_t,\; x^{(i)}_{t+1},\; l_i,\; \mathbf{1}_{[t = T_i - 1]} \cdot \hat{r}_i ) \;\big|\; \tau_i \in \sD,\;\; t = 0, \dots, T_i - 1 \,\big\}.
\label{eq:augmented_dataset}
\end{equation}

RoboFQE fits an action-value function $Q_\theta: \sX \times \sA \times \sL \to \mathbb{R}$ on $\sD'$, where $\sX$ is the space of policy inputs $x_t = f(h_t)$ and $\theta$ denotes all trainable parameters of $Q_\theta$, including the visual encoder. The value function is trained by minimizing the following squared-error loss:
\begin{equation}
L_{\mathrm{FQE}}(\theta)
\;=\;
\E_{(x^{(i)}_t,\,a^{(i)}_t,\,x^{(i)}_{t+1},\,l_i) \sim \sD'}
\!\left[
\left(
Q_\theta\!\left(x^{(i)}_t, a^{(i)}_t, l_i\right)
-
y^{(i)}_t
\right)^2
\right].
\label{eq:fqe_loss}
\end{equation}
For each transition $(x^{(i)}_t, a^{(i)}_t, x^{(i)}_{t+1}, l_i) \in \sD'$, the FQE target is:
\begin{equation}
y^{(i)}_t \;=\;
\begin{cases}
\gamma\, \E_{a' \sim \pi(\cdot \mid x^{(i)}_{t+1}, l_i)}
\!\left[
Q_{\bar{\theta}}\!\left(x^{(i)}_{t+1}, a', l_i\right)
\right],
& \text{if } t < T_i - 1, \\
\hat{r}_i,
& \text{if } t = T_i - 1,
\end{cases}
\label{eq:fqe_target}
\end{equation}
where $\bar{\theta}$ denotes a periodically synchronized copy of $\theta$ (target network).
For deterministic policies, the expectation over $a'$ reduces to evaluating $Q_{\bar{\theta}}$ at $a' = \pi(x^{(i)}_{t+1}, l_i)$; for stochastic policies, it is estimated using Monte Carlo action samples. Fitting these targets requires $\sD'$ to cover the inputs and actions at which they query $Q_{\bar{\theta}}$. Because $x_t$ may discard information from the full history, the Bellman recursion that motivates Eq.~\eqref{eq:fqe_target} also requires that $x_t$ be sufficient for predicting future returns under $\pi$~\citep{uehara2023future}, i.e., $x_t$ must be a Markov state representation. When this condition does not hold, distinct latent situations with different future returns may be represented by the same $x_t$, forcing any value function conditioned only on $x_t$ to assign them the same prediction.
A separate source of error arises from the quality of the visual representation: the action-value function must operate on features derived from high-dimensional observations, and the quality of these features affects value-estimation accuracy.

After offline fitting, RoboFQE evaluates policy from initial inputs without generating trajectories:
\begin{equation}
\widehat{V}_l^\pi
\;=\;
\frac{1}{|\mathcal{I}_l|}
\sum_{i \in \mathcal{I}_l}
\E_{a \sim \pi(\cdot \mid x^{(i)}_0, l)}
\!\left[
Q_\theta\!\left(x^{(i)}_0, a, l\right)
\right],
\qquad
\mathcal{I}_l = \{\, i : l_i = l \,\}.
\label{eq:value_estimate}
\end{equation}
This estimate uses the empirical distribution of initial inputs for task $l$ in $\sD$; if the deployment initial-state distribution differs from the offline dataset, this mismatch becomes an additional source of evaluation error.

\subsection{Instantiation for Visual Robot Policies}
\label{sec:robofqe_instantiation}

FQE cannot be applied to visual robot policies without resolving two issues: encoding high-dimensional observations, and obtaining rewards when success labels are absent. RoboFQE addresses both through its choice of visual representation and success detector.
The action-value function processes image observations through DINOv3~\citep{simeoni2025dinov3}, a self-supervised vision foundation model. Non-visual inputs (proprioceptive states, gripper status) and the continuous action are projected through learned embedding layers and concatenated with the visual features. The language instruction is encoded via a pretrained text encoder and fused with the observation--action representation to produce the scalar action-value estimate $Q_\theta(x_t, a_t, l)$. When ground-truth terminal success labels are unavailable, terminal success is inferred using the Large Reward Model~\citep[LRM;][]{wu2026large}, a vision-language model, used off the shelf or adapted on demonstration data, that produces a binary success judgment from the terminal observation and task instruction. The success detector labels $\sD$ as a preprocessing step before FQE training begins.
The pretrained representation supports value learning from high-dimensional observations, while the success detector supplies terminal success labels in place of ground-truth annotation. Together, these two adaptations make value-based OPE applicable to visuomotor policy evaluation without altering the FQE objective itself.

\section{Experiments}
\label{sec:experiments}

We evaluate RoboFQE on two benchmarks chosen for complementary properties. LIBERO~\citep{liu2023libero} provides a comparison of how estimation error grows with horizon. CALVIN~\citep{mees2022calvin} adds a longer-horizon sequential regime in which value estimates correspond to the number of tasks completed. A real-world study on an SO-101 arm then tests RoboFQE with its success detector applied zero-shot.

\subsection{Setup}
\label{sec:experimental_setup}

\paragraph{Benchmark and data.}
The four LIBERO~\citep{liu2023libero} suites are Spatial ($H{=}220$), Object ($H{=}280$), Goal ($H{=}300$), and Long ($H{=}520$), each containing 10 language-conditioned manipulation tasks with sparse binary terminal reward. This progression in horizon length enables controlled comparison of how estimation accuracy degrades as the evaluation horizon grows. Terminal success labels for augmented rollouts are annotated by LRM~\citep{wu2026large} fine-tuned on the official LIBERO demonstration data, achieving 97.4\% accuracy against simulator ground-truth on the augmented rollouts. The offline dataset $\sD$ comprises 50 human demonstrations and 25 additional rollouts per task, totaling 750 trajectories per suite across 10 tasks (102k, 121k, 114k, and 232k transitions for Spatial, Object, Goal, and Long, respectively). The additional rollouts are collected from partially trained behavioral cloning policies using the ResNet-T architecture and the official LIBERO training pipeline.

On CALVIN~\citep{mees2022calvin}, policies are trained on environments A--C and evaluated on the held-out environment D, following the ABC$\to$D split. All methods use the same 20 fixed task chains, each a sequence of five subtasks executed consecutively in D. The offline dataset is drawn from the A--C data, which, unlike the LIBERO demonstrations, contains both successful and failed trajectories; no additional rollout augmentation is performed.

\paragraph{Target policies and ground truth.}
On LIBERO, the target policies use the ResNet-T architecture (ResNet encoder with transformer temporal decoder) provided by LIBERO~\citep{liu2023libero}, trained via behavioral cloning on the benchmark's demonstration data. For absolute value estimation, a single policy is evaluated per suite across 5 seeds. For ranking, four policies from the same training run, checkpointed at varying stages of training, are compared per suite, producing a range of separated success rates. On CALVIN, the target policy is X-VLA~\citep{zheng2025x}, using the official checkpoint fine-tuned on the ABC split. Ground-truth returns are computed by Monte Carlo rollouts in the simulator: on LIBERO with $\gamma = 0.99$, averaging $\gamma^{T_i-1} \cdot R_l(s_{T_i})$ over 200 episodes per policy; on CALVIN with $\gamma{=}1$, executing the policy over the same 20 fixed task sequences, each consisting of five subtasks to be performed sequentially in environment D.
 
\paragraph{Baselines.}
\textsc{Simulator Rollout} executes the target policy directly in the LIBERO simulator, producing the empirical success rate (SR).
WorldGym~\citep{quevedo2026worldgym} rolls out the target policy in an action-conditioned video diffusion model and classifies terminal success via a VLM. Cosmos 2.5-AC adapts the pretrained Cosmos Video2World diffusion transformer into an action-conditional world model following \citet{tseng2025scalable} and applies the same VLM-based success classification. Ctrl-World~\citep{guo2026ctrl}, evaluated on the four LIBERO suites, is a multi-view world model that maintains long-horizon consistency through pose-conditioned memory retrieval and uses the same success classification. All three world models generate rollouts chunk by chunk, conditioning each chunk on previously generated frames and the policy's actions, and are post-trained on $\sD$ to adapt to the LIBERO observations.
RoboFQE uses a DINOv3 encoder fine-tuned on the offline dataset $\sD$ via the FQE loss with LRM annotated terminal labels, instantiated as in Section~\ref{sec:robofqe_instantiation}.
Value estimation accuracy is measured by the absolute relative error (ARE) and mean squared error (MSE) of the estimated discounted return, with lower values indicating more accurate estimates. Ranking accuracy is measured by the Pearson correlation $r$ and the mean maximum rank violation (MMRV), with higher $r$ and lower MMRV indicating more faithful rankings.
 
\subsection{Evaluation on LIBERO}
\label{sec:main_results}
 
Table~\ref{tab:combined_results} reports value estimation accuracy and policy ranking accuracy across the four LIBERO suites, following the evaluation protocol of \citet{li24simpler}, \citet{quevedo2026worldgym}, and \citet{tseng2025scalable}. Ranking metrics are reported against both ground-truth discounted return $V^\pi$ and SR, because these two targets can induce different policy orderings: $V^\pi$ weights terminal success by $\gamma^{T-1}$, penalizing policies that succeed only after many steps, whereas SR treats all successes equally. Reporting both value estimation error and ranking metrics distinguishes accurate value estimation from the ranking-only evaluation that prior work has focused on.

\begin{table}[t]
\centering
\setlength{\belowcaptionskip}{0pt} 
\caption{Value estimation and policy ranking on LIBERO. All uncertainties are 95\% bootstrapped confidence intervals. Ground-truth per suite ($\gamma{=}0.99$): Spatial ($V^\pi{=}0.237$, SR${=}0.84$), Object ($V^\pi{=}0.166$, SR${=}0.80$), Goal ($V^\pi{=}0.246$, SR${=}0.77$), Long ($V^\pi{=}0.029$, SR${=}0.43$).}
\label{tab:combined_results}
\vspace{0.6em}
\small
\setlength{\tabcolsep}{4pt}
\renewcommand{\arraystretch}{0.95}
\newcommand{\ci}[1]{{\scriptsize$\pm#1$}}
\begin{tabular}{@{}llcccccc@{}}
\toprule
 & & & & \multicolumn{2}{c}{vs $V^\pi$} & \multicolumn{2}{c}{vs SR} \\
\cmidrule(lr){5-6} \cmidrule(lr){7-8}
Method & Suite & ARE (\%) $\downarrow$ & MSE ($\times 10^{-5}$) $\downarrow$ & $r$ $\uparrow$ & MMRV $\downarrow$ & $r$ $\uparrow$ & MMRV $\downarrow$ \\
\midrule
\multirow{4}{*}{RoboFQE}
 & Spatial & $\mathbf{3.40}$~\ci{0.08}   & $\mathbf{6.50}$~\ci{0.31} & $\mathbf{+0.96}$ & $\mathbf{0.03}$ & $\mathbf{+0.96}$ & $0.11$ \\
 & Object  & $\mathbf{4.56}$~\ci{0.17}   & $\mathbf{5.74}$~\ci{0.44} & $\mathbf{+0.91}$ & $\mathbf{0.02}$ & $\mathbf{+0.90}$ & $0.10$ \\
 & Goal    & $\mathbf{2.34}$~\ci{0.19}   & $\mathbf{3.38}$~\ci{0.59} & $\mathbf{+0.90}$ & $\mathbf{0.05}$ & $\mathbf{+0.90}$ & $0.13$ \\
 & Long    & $\mathbf{5.63}$~\ci{2.71}   & $\mathbf{0.42}$~\ci{0.38} & $\mathbf{+0.88}$ & $\mathbf{0.07}$ & $\mathbf{+0.89}$ & $\mathbf{0.15}$ \\
\midrule
\multirow{4}{*}{RoboFQE-scratch}
 & Spatial & $9.35$~\ci{1.88}   & $46.7$~\ci{9.3}  & $+0.84$ & $0.09$ & $+0.84$ & $0.12$ \\
 & Object  & $13.89$~\ci{4.79}  & $73.2$~\ci{24.9} & $+0.83$ & $0.11$ & $+0.83$ & $0.14$ \\
 & Goal    & $5.82$~\ci{1.72}   & $26.2$~\ci{7.9}  & $+0.83$ & $0.09$ & $+0.83$ & $0.19$ \\
 & Long    & $32.09$~\ci{8.83}  & $9.7$~\ci{2.7}   & $+0.72$ & $0.17$ & $+0.72$ & $0.27$ \\
\midrule
\multirow{4}{*}{WorldGym}
 & Spatial & $24.20$~\ci{4.72}  & $337.49$~\ci{104.26}  & $+0.76$ & $0.18$ & $+0.78$ & $0.16$ \\
 & Object  & $22.21$~\ci{5.66}  & $136.57$~\ci{64.89}   & $+0.70$ & $0.20$ & $+0.77$ & $0.21$ \\
 & Goal    & $50.04$~\ci{7.47}  & $1572.30$~\ci{452.08} & $+0.62$ & $0.24$ & $+0.64$ & $0.22$ \\
 & Long    & $69.18$~\ci{10.63} & $39.75$~\ci{12.44}    & $-0.24$ & $0.68$ & $-0.35$ & $0.70$ \\
\midrule
\multirow{4}{*}{Cosmos 2.5-AC}
 & Spatial & $8.44$~\ci{0.19}   & $40.33$~\ci{2.09}  & $+0.81$ & $0.10$ & $+0.82$ & $0.10$ \\
 & Object  & $7.55$~\ci{0.83}   & $15.77$~\ci{4.30}  & $+0.82$ & $0.09$ & $+0.82$ & $\mathbf{0.09}$ \\
 & Goal    & $8.03$~\ci{2.37}   & $39.27$~\ci{19.89} & $+0.84$ & $0.09$ & $+0.84$ & $\mathbf{0.10}$ \\
 & Long    & $54.37$~\ci{14.89} & $21.65$~\ci{10.91} & $+0.56$ & $0.26$ & $+0.60$ & $0.23$ \\
\midrule
\multirow{4}{*}{Ctrl-World}
 & Spatial & $9.13$~\ci{1.04}   & $46.52$~\ci{2.15}   & $+0.81$  & $0.08$ & $+0.81$  & $\mathbf{+0.09}$ \\
 & Object  & $8.17$~\ci{1.47}   & $18.45$~\ci{4.84}   & $+0.80$  & $0.07$ & $+0.82$  & $+0.10$ \\
 & Goal    & $8.71$~\ci{3.90}   & $46.09$~\ci{11.42}  & $+0.81$ & $0.09$ & $+0.82$ & $+0.11$ \\
 & Long    & $58.75$~\ci{7.22}  & $26.53$~\ci{23.42}  & $+0.60$ & $0.22$ & $+0.64$ & $0.22$ \\
\bottomrule
\end{tabular}
\vspace{-0.6em}
\end{table}

\paragraph{Value estimation accuracy.}
RoboFQE achieves the lowest ARE across all four suites. Cosmos 2.5-AC is competitive on the shorter-horizon suites but degrades sharply with horizon: on Long, all three world-model baselines exceed 50\% ARE while RoboFQE remains below 6\%. This degradation appears across all three world-model baselines despite their differing architectures and scales, including Ctrl-World's memory retrieval designed for long-horizon consistency, suggesting it is not specific to a single evaluator and is consistent with error accumulation in autoregressive rollout generation~\citep{sutton1996model,xiao2019learning}.

\paragraph{Policy ranking.}
Against $V^\pi$, RoboFQE achieves the highest Pearson $r$ and the lowest MMRV on every suite. WorldGym's ranking on Long inverts entirely ($r = -0.24$), while Cosmos 2.5-AC and Ctrl-World retain positive correlation but with higher rank violation. Against SR, Cosmos 2.5-AC attains the lowest MMRV on the three shorter-horizon suites, but RoboFQE recovers it on Long, making RoboFQE the only method whose ranking accuracy remains stable as horizon increases. Conversely, strong ranking does not imply accurate value estimates: Cosmos 2.5-AC achieves $r \geq 0.81$ on Spatial through Goal but with ARE well above RoboFQE's.

\paragraph{Effect of large-scale visual pretraining.}
To isolate the contribution of the pretrained encoder, RoboFQE-scratch replaces DINOv3 with ResNet-18 backbones trained from scratch, with all other components identical. RoboFQE-scratch increases ARE on every suite, most severely on Long, and ranking follows the same pattern.

\subsection{Evaluation on CALVIN}
\label{sec:calvin_results}

\begin{table}[b]
\centering
\caption{Value estimation on CALVIN (ABC$\to$D, $\gamma{=}1$). Target policy: X-VLA~\citep{zheng2025x}. SR@$k$: success rate of completing at least $k$ consecutive tasks. Avg.\ Len.: average number of consecutively completed tasks.}
\label{tab:calvin_results}
\vspace{0.6em}
\small
\begin{tabular}{@{}lcccccc@{}}
\toprule
Estimator & SR@1 & SR@2 & SR@3 & SR@4 & SR@5 & Avg.\ Len. \\
\midrule
Ground Truth  & 0.96  & 0.91  & 0.87  & 0.82  & 0.76  & 4.32   \\
RoboFQE       & 0.95  & 0.82  & 0.69  & 0.61  & 0.54  & 3.61   \\
WorldGym      & 0.75  & 0.24  & 0.05  & 0.00  & 0.00  & 1.04   \\
Cosmos 2.5-AC & 0.98  & 0.42  & 0.08  & 0.00  & 0.00  & 1.48   \\
Ctrl-World    & 0.96  & 0.51  & 0.10  & 0.01  & 0.00  & 1.59   \\
\bottomrule
\end{tabular}
\vspace{-0.6em}
\end{table}

Table~\ref{tab:calvin_results} reports value estimation on CALVIN for the X-VLA policy~\citep{zheng2025x}. RoboFQE matches the ground-truth first-subtask rate ($0.95$ versus $0.96$) and tracks the ground-truth rates along the chain, underestimating more at greater depth and arriving at an estimated average length within $17\%$ of ground truth. Both world-model baselines recover the first-subtask rate but collapse from the second subtask onward.

Both rollout-based world-model evaluators generate the entire chain, so generation error accumulates across subtask transitions and degrades the frames on which each subtask is initialized and scored. Cosmos 2.5-AC's stronger generator raises the first-subtask rate but does not avoid this accumulation. RoboFQE's underestimate grows with chain depth: the entry inputs for a deeper subtask, reached only after all preceding subtasks succeed, are sparser in the A--C data.

\needspace{5\baselineskip}
\subsection{Real-World Evaluation}
\label{sec:realworld}

\begin{wrapfigure}{r}{0.36\textwidth}
\vspace{-1.6em}
\centering
\includegraphics[width=\linewidth]{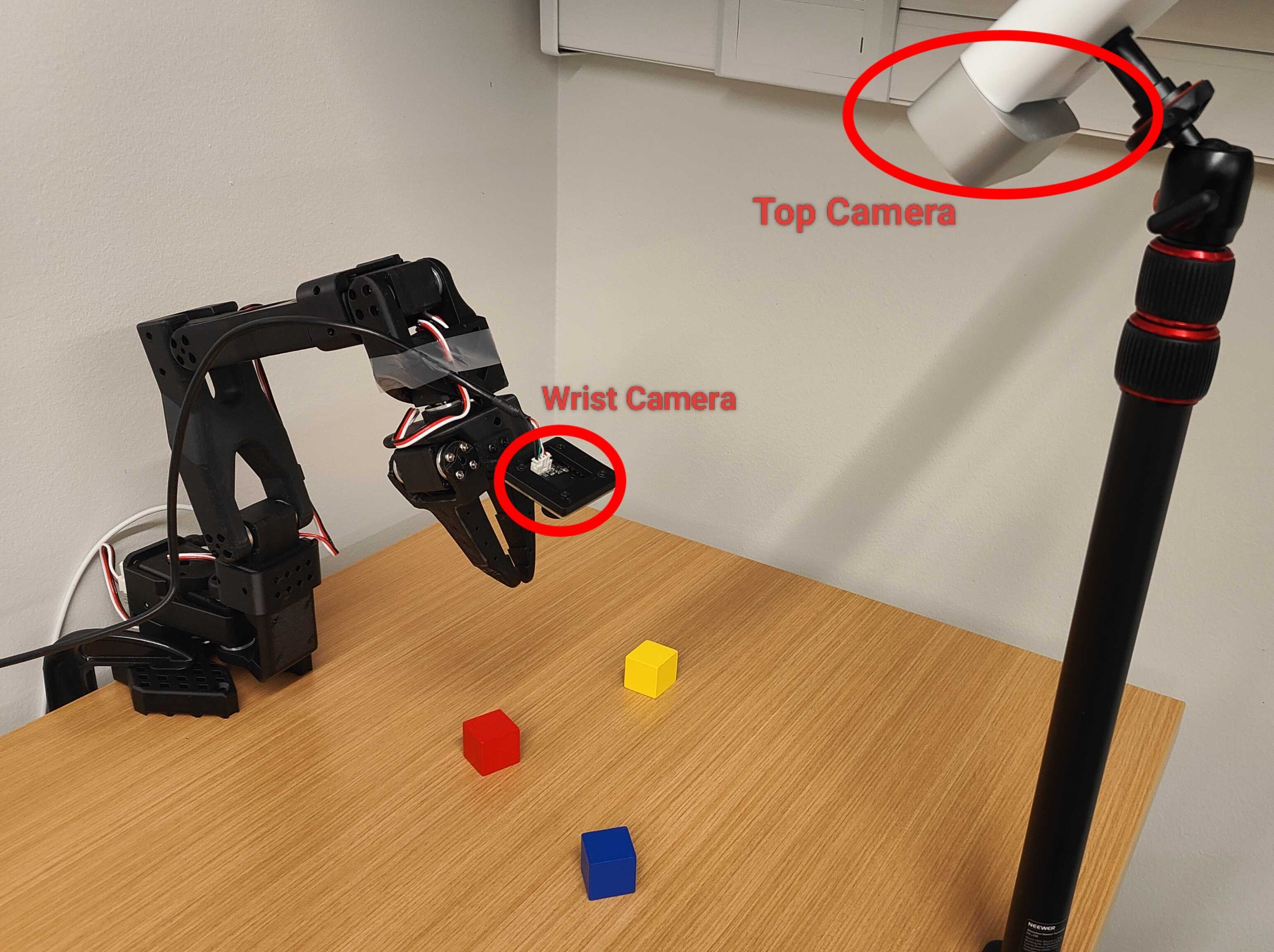}
\vspace{-0.4em}
\caption{Real-world setup: an SO-101 arm with a top camera and a wrist camera stacks colored blocks.}
\label{fig:real_world_setup}
\vspace{-1.4em}
\end{wrapfigure}

The real-world study (Figure~\ref{fig:real_world_setup}) uses an SO-101 arm with a top camera and a wrist camera on three block-stacking tasks, which share the same red, yellow, and blue blocks and differ only in the source and target block; Appendix~\ref{app:realworld} gives the full setup. The offline dataset combines 150 teleoperated demonstrations, collected under five initial-position settings per task, with 100 rollouts of a behavioral cloning data-collection policy. The target policies are SmolVLA~\citep{shukor2025smolvla}, $\pi_{0.5}$~\citep{black2025pi05}, and a multitask behavioral cloning policy, none of which contributes trajectories to the offline dataset; over 100 robot episodes per task with $\gamma{=}1$, their ground-truth success rates are 58.5\%, 81.0\%, and 33.0\%, respectively. LRM is applied zero-shot and labels terminal success with 83.3\% accuracy against manual labels, compared with 97.4\% after fine-tuning on LIBERO. WorldGym and Cosmos 2.5-AC are post-trained on the same offline dataset.

\looseness=-1 As Table~\ref{tab:realworld_results} shows, RoboFQE achieves the lowest ARE and the highest correlation despite the noisier labels. Both world-model evaluators incur higher estimation error, and Cosmos 2.5-AC again combines moderate ranking accuracy with high estimation error, as on LIBERO. Real-world trajectories average approximately 500 steps, comparable to LIBERO-Long, where world-model evaluators degrade most.

\begin{table}[t]
\centering
\setlength{\belowcaptionskip}{0pt} 
\caption{Value estimation and policy ranking on real-world block-stacking tasks, aggregated across three tasks. With $\gamma{=}1$, $V^\pi$ equals SR, so ranking metrics against either are identical. ARE and MSE: mean $\pm$ 95\% $t$-CI half-width over three seeds.}
\label{tab:realworld_results}
\vspace{0.6em}
\small
\setlength{\tabcolsep}{4pt}
\begin{tabular}{@{}lcccc@{}}
\toprule
Method & ARE (\%) $\downarrow$ & MSE ($\times 10^{-2}$) $\downarrow$ & $r$ $\uparrow$ & MMRV $\downarrow$ \\
\midrule
RoboFQE       & $\mathbf{7.5} \pm 5.0$  & $\mathbf{0.22} \pm 0.30$ & $\mathbf{+0.90}$ & $\mathbf{0.11}$ \\
WorldGym      & $17.8 \pm 10.9$ & $1.24 \pm 1.52$ & $+0.44$ & $0.40$ \\
Cosmos 2.5-AC & $24.5 \pm 17.1$ & $2.35 \pm 3.28$ & $+0.68$ & $0.28$ \\
Ctrl-World    & $10.3 \pm 10.2$ & $1.07 \pm 1.57$ & $+0.49$ & $0.19$ \\
\bottomrule
\end{tabular}
\vspace{-0.6em}
\end{table}

\section{Conclusion and Limitations}
RoboFQE demonstrates that fitted Q-evaluation with a pretrained vision encoder produces more accurate return estimates than video world-model evaluators, especially as the manipulation horizon increases. By reporting absolute error on value estimation alongside ranking metrics, RoboFQE reveals that strong rank correlation does not imply calibrated estimates, a distinction that matters when deployment decisions depend on absolute performance thresholds.

We note three limitations of our study. First, value-based OPE accuracy depends on the offline dataset's coverage of state-action regions visited by the target policy; quantifying this dependence under visual observations remains an open question.
Our study does not establish whether RoboFQE can reliably rank policies on tasks absent from the offline dataset, or how its generalization compares with that of large-scale pretrained world-model evaluators.

Second, when ground-truth rewards are unavailable, errors in the learned success detector's annotations can lead to inaccurate value estimates, and vision-language reward models are expected to degrade on out-of-distribution real-world scenes.
Third, RoboFQE conditions the Q-function on the policy input rather than full history or a Markov state, which could lead to less accurate prediction in partially observable environments.


\clearpage
\acknowledgments{This work took place in the Prediction and Action Lab (PAL) at the University of Wisconsin–Madison. PAL research is supported by NSF (IIS-2410981, IIS-2551198) and the Wisconsin Alumni Research Foundation.}


\bibliography{ref}  

@article{tseng2025scalable,
  title={Scalable policy evaluation with video world models},
  author={Tseng, Wei-Cheng and Gu, Jinwei and Zhang, Qinsheng and Mao, Hanzi and Liu, Ming-Yu and Shkurti, Florian and Yen-Chen, Lin},
  journal={arXiv preprint arXiv:2511.11520},
  year={2025}
}

@article{kaelbling1998planning,
  title={Planning and acting in partially observable stochastic domains},
  author={Kaelbling, Leslie Pack and Littman, Michael L and Cassandra, Anthony R},
  journal={Artificial intelligence},
  volume={101},
  number={1-2},
  pages={99--134},
  year={1998},
  publisher={Elsevier}
}

@article{zhou2025autoeval,
  title={Autoeval: Autonomous evaluation of generalist robot manipulation policies in the real world},
  author={Zhou, Zhiyuan and Atreya, Pranav and Tan, You Liang and Pertsch, Karl and Levine, Sergey},
  journal={arXiv preprint arXiv:2503.24278},
  year={2025}
}

@article{atreya2025roboarena,
  author = {Pranav Atreya and Karl Pertsch and Tony Lee and Moo Jin Kim and Arhan Jain and Artur Kuramshin and Clemens Eppner and Cyrus Neary and Edward Hu and Fabio Ramos and Jonathan Tremblay and Kanav Arora and Kirsty Ellis and Luca Macesanu and Matthew Leonard and Meedeum Cho and Ozgur Aslan and Shivin Dass and Jie Wang and Xingfang Yuan and Xuning Yang and Abhishek Gupta and Dinesh Jayaraman and Glen Berseth and Kostas Daniilidis and Roberto Martin-Martin and Youngwoon Lee and Percy Liang and Chelsea Finn and Sergey Levine},
  journal = {arXiv},
  title = {RoboArena: Distributed Real-World Evaluation of Generalist Robot Policies},
  year = {2025},
}

@article{li24simpler,
    title={Evaluating Real-World Robot Manipulation Policies in Simulation},
    author={Xuanlin Li and Kyle Hsu and Jiayuan Gu and Karl Pertsch and Oier Mees and Homer Rich Walke and Chuyuan Fu and Ishikaa Lunawat and Isabel Sieh and Sean Kirmani and Sergey Levine and Jiajun Wu and Chelsea Finn and Hao Su and Quan Vuong and Ted Xiao},
    journal = {arXiv preprint arXiv:2405.05941},
    year={2024},
}

@article{badithela2025reliable,
  title={Reliable and scalable robot policy evaluation with imperfect simulators},
  author={Badithela, Apurva and Snyder, David and Zha, Lihan and Mikhail, Joseph and O'Kelly, Matthew and Dixit, Anushri and Majumdar, Anirudha},
  journal={arXiv preprint arXiv:2510.04354},
  year={2025}
}

@article{angelopoulos2023prediction,
  title={Prediction-powered inference},
  author={Angelopoulos, Anastasios N and Bates, Stephen and Fannjiang, Clara and Jordan, Michael I and Zrnic, Tijana},
  journal={Science},
  volume={382},
  number={6671},
  pages={669--674},
  year={2023},
  publisher={American Association for the Advancement of Science}
}

@inproceedings{
quevedo2026worldgym,
title={WorldGym: World Model as An Environment for Policy Evaluation},
author={Julian Hector Quevedo and Ansh Kumar Sharma and Yixiang Sun and Varad Suryavanshi and Percy Liang and Sherry Yang},
booktitle={The Fourteenth International Conference on Learning Representations},
year={2026},
url={https://openreview.net/forum?id=hidBHy1CAw}
}

@misc{veorobotics2025,
      title={Evaluating Gemini Robotics Policies in a Veo World Simulator}, 
      author={Gemini Robotics Team and Krzysztof Choromanski and Coline Devin and Yilun Du and Debidatta Dwibedi and Ruiqi Gao and Abhishek Jindal and Thomas Kipf and Sean Kirmani and Isabel Leal and Fangchen Liu and Anirudha Majumdar and Andrew Marmon and Carolina Parada and Yulia Rubanova and Dhruv Shah and Vikas Sindhwani and Jie Tan and Fei Xia and Ted Xiao and Sherry Yang and Wenhao Yu and Allan Zhou},
      year={2025},
      eprint={2512.10675},
      archivePrefix={arXiv},
      primaryClass={cs.RO},
      url={https://arxiv.org/abs/2512.10675}, 
}

@inproceedings{sutton1996model,
  title={Model-based reinforcement learning with an approximate, learned model},
  author={Sutton, Leonid Kuvayev Rich},
  booktitle={Proceedings of the ninth Yale workshop on adaptive and learning systems},
  volume={1996},
  pages={101--105},
  year={1996}
}

@article{xiao2019learning,
  title={Learning to combat compounding-error in model-based reinforcement learning},
  author={Xiao, Chenjun and Wu, Yifan and Ma, Chen and Schuurmans, Dale and M{\"u}ller, Martin},
  journal={arXiv preprint arXiv:1912.11206},
  year={2019}
}

@InProceedings{pmlr-v97-le19a,
  title = 	 {Batch Policy Learning under Constraints},
  author =       {Le, Hoang and Voloshin, Cameron and Yue, Yisong},
  booktitle = 	 {Proceedings of the 36th International Conference on Machine Learning},
  pages = 	 {3703--3712},
  year = 	 {2019},
  editor = 	 {Chaudhuri, Kamalika and Salakhutdinov, Ruslan},
  volume = 	 {97},
  series = 	 {Proceedings of Machine Learning Research},
  month = 	 {09--15 Jun},
  publisher =    {PMLR},
  url = 	 {https://proceedings.mlr.press/v97/le19a.html}
}

@article{fu2021benchmarks,
  title={Benchmarks for deep off-policy evaluation},
  author={Fu, Justin and Norouzi, Mohammad and Nachum, Ofir and Tucker, George and Wang, Ziyu and Novikov, Alexander and Yang, Mengjiao and Zhang, Michael R and Chen, Yutian and Kumar, Aviral and others},
  journal={arXiv preprint arXiv:2103.16596},
  year={2021}
}

@article{nachum2019dualdice,
  title={DualDICE: Behavior-Agnostic Estimation of Discounted Stationary
  Distribution Corrections},
  author={Nachum, Ofir and Chow, Yinlam and Dai, Bo and Li, Lihong},
  journal={NeurIPS},
  year={2019}
}

@article{yang2020off,
  title={Off-policy evaluation via the regularized lagrangian},
  author={Yang, Mengjiao and Nachum, Ofir and Dai, Bo and Li, Lihong and Schuurmans, Dale},
  journal={Advances in Neural Information Processing Systems},
  volume={33},
  pages={6551--6561},
  year={2020}
}

@InProceedings{pmlr-v48-jiang16,
  title = 	 {Doubly Robust Off-policy Value Evaluation for Reinforcement Learning},
  author = 	 {Jiang, Nan and Li, Lihong},
  booktitle = 	 {Proceedings of The 33rd International Conference on Machine Learning},
  pages = 	 {652--661},
  year = 	 {2016},
  editor = 	 {Balcan, Maria Florina and Weinberger, Kilian Q.},
  volume = 	 {48},
  series = 	 {Proceedings of Machine Learning Research},
  address = 	 {New York, New York, USA},
  month = 	 {20--22 Jun},
  publisher =    {PMLR},
  url = 	 {https://proceedings.mlr.press/v48/jiang16.html}
}

@InProceedings{pmlr-v48-thomasa16,
  title = 	 {Data-Efficient Off-Policy Policy Evaluation for Reinforcement Learning},
  author = 	 {Thomas, Philip and Brunskill, Emma},
  booktitle = 	 {Proceedings of The 33rd International Conference on Machine Learning},
  pages = 	 {2139--2148},
  year = 	 {2016},
  editor = 	 {Balcan, Maria Florina and Weinberger, Kilian Q.},
  volume = 	 {48},
  series = 	 {Proceedings of Machine Learning Research},
  address = 	 {New York, New York, USA},
  month = 	 {20--22 Jun},
  publisher =    {PMLR},
  url = 	 {https://proceedings.mlr.press/v48/thomasa16.html}
}

@inproceedings{
  voloshin2021empirical,
  title={Empirical Study of Off-Policy Policy Evaluation for Reinforcement Learning},
  author={Cameron Voloshin and Hoang Minh Le and Nan Jiang and Yisong Yue},
  booktitle={Thirty-fifth Conference on Neural Information Processing Systems Datasets and Benchmarks Track (Round 1)},
  year={2021},
  url={https://openreview.net/forum?id=IsK8iKbL-I}
}

@inproceedings{pavse2023state,
  title={State-Action Similarity-Based Representations for Off-Policy Evaluation},
  author={Pavse, Brahma S and Hanna, Josiah P},
  booktitle={Advances in Neural Information Processing Systems},
  volume={36},
  year={2023}
}

@inproceedings{pavse2025stable,
  title={Stable Offline Value Function Learning with Bisimulation-based Representations},
  author={Pavse, Brahma S and Chen, Yudong and Xie, Qiaomin and Hanna, Josiah P},
  booktitle={Proceedings of the 42nd International Conference on Machine Learning},
  year={2025}
}

@article{ernst2005tree,
  author  = {Damien  Ernst and Pierre  Geurts and Louis  Wehenkel},
  title   = {Tree-Based Batch Mode Reinforcement Learning},
  journal = {Journal of Machine Learning Research},
  year    = {2005},
  volume  = {6},
  number  = {18},
  pages   = {503--556},
  url     = {http://jmlr.org/papers/v6/ernst05a.html}
}

@misc{black2026pi0visionlanguageactionflowmodel,
      title={$\pi_0$: A Vision-Language-Action Flow Model for General Robot Control}, 
      author={Kevin Black and Noah Brown and Danny Driess and Adnan Esmail and Michael Equi and Chelsea Finn and Niccolo Fusai and Lachy Groom and Karol Hausman and Brian Ichter and Szymon Jakubczak and Tim Jones and Liyiming Ke and Sergey Levine and Adrian Li-Bell and Mohith Mothukuri and Suraj Nair and Karl Pertsch and Lucy Xiaoyang Shi and James Tanner and Quan Vuong and Anna Walling and Haohuan Wang and Ury Zhilinsky},
      year={2026},
      eprint={2410.24164},
      archivePrefix={arXiv},
      primaryClass={cs.LG},
      url={https://arxiv.org/abs/2410.24164}, 
}

@article{team2024octo,
  title={Octo: An open-source generalist robot policy},
  author={Team, Octo Model and Ghosh, Dibya and Walke, Homer and Pertsch, Karl and Black, Kevin and Mees, Oier and Dasari, Sudeep and Hejna, Joey and Kreiman, Tobias and Xu, Charles and others},
  journal={arXiv preprint arXiv:2405.12213},
  year={2024}
}

@article{chi2025diffusion,
  title={Diffusion policy: Visuomotor policy learning via action diffusion},
  author={Chi, Cheng and Xu, Zhenjia and Feng, Siyuan and Cousineau, Eric and Du, Yilun and Burchfiel, Benjamin and Tedrake, Russ and Song, Shuran},
  journal={The International Journal of Robotics Research},
  volume={44},
  number={10-11},
  pages={1684--1704},
  year={2025},
  publisher={Sage Publications Sage UK: London, England}
}

@article{bjorck2025gr00t,
  title={Gr00t n1: An open foundation model for generalist humanoid robots},
  author={Bjorck, Johan and Casta{\~n}eda, Fernando and Cherniadev, Nikita and Da, Xingye and Ding, Runyu and Fan, Linxi and Fang, Yu and Fox, Dieter and Hu, Fengyuan and Huang, Spencer and others},
  journal={arXiv preprint arXiv:2503.14734},
  year={2025}
}

@inproceedings{liu2025rdt,
  title={Rdt-1b: a diffusion foundation model for bimanual manipulation},
  author={Liu, Songming and Wu, Lingxuan and Li, Bangguo and Tan, Hengkai and Chen, Huayu and Wang, Zhengyi and Xu, Ke and Su, Hang and Zhu, Jun},
  booktitle={International Conference on Learning Representations},
  volume={2025},
  pages={29982--30009},
  year={2025}
}

@article{shukor2025smolvla,
  title={Smolvla: A vision-language-action model for affordable and efficient robotics},
  author={Shukor, Mustafa and Aubakirova, Dana and Capuano, Francesco and Kooijmans, Pepijn and Palma, Steven and Zouitine, Adil and Aractingi, Michel and Pascal, Caroline and Russi, Martino and Marafioti, Andres and others},
  journal={arXiv preprint arXiv:2506.01844},
  year={2025}
}

@article{kim2024openvla,
  title={Openvla: An open-source vision-language-action model},
  author={Kim, Moo Jin and Pertsch, Karl and Karamcheti, Siddharth and Xiao, Ted and Balakrishna, Ashwin and Nair, Suraj and Rafailov, Rafael and Foster, Ethan and Lam, Grace and Sanketi, Pannag and others},
  journal={arXiv preprint arXiv:2406.09246},
  year={2024}
}

@article{liu2023libero,
  title={Libero: Benchmarking knowledge transfer for lifelong robot learning},
  author={Liu, Bo and Zhu, Yifeng and Gao, Chongkai and Feng, Yihao and Liu, Qiang and Zhu, Yuke and Stone, Peter},
  journal={Advances in Neural Information Processing Systems},
  volume={36},
  pages={44776--44791},
  year={2023}
}

@misc{jain2025polarisscalablerealtosimevaluations,
      title={PolaRiS: Scalable Real-to-Sim Evaluations for Generalist Robot Policies}, 
      author={Arhan Jain and Mingtong Zhang and Kanav Arora and William Chen and Marcel Torne and Muhammad Zubair Irshad and Sergey Zakharov and Yue Wang and Sergey Levine and Chelsea Finn and Wei-Chiu Ma and Dhruv Shah and Abhishek Gupta and Karl Pertsch},
      year={2025},
      eprint={2512.16881},
      archivePrefix={arXiv},
      primaryClass={cs.RO},
      url={https://arxiv.org/abs/2512.16881}, 
}

@article{simeoni2025dinov3,
  title={Dinov3},
  author={Sim{\'e}oni, Oriane and Vo, Huy V and Seitzer, Maximilian and Baldassarre, Federico and Oquab, Maxime and Jose, Cijo and Khalidov, Vasil and Szafraniec, Marc and Yi, Seungeun and Ramamonjisoa, Micha{\"e}l and others},
  journal={arXiv preprint arXiv:2508.10104},
  year={2025}
}

@article{wu2026large,
  title={Large Reward Models: Generalizable Online Robot Reward Generation with Vision-Language Models},
  author={Wu, Yanru and Yuan, Weiduo and Qi, Ang and Guizilini, Vitor and Mao, Jiageng and Wang, Yue},
  journal={arXiv preprint arXiv:2603.16065},
  year={2026}
}

@article{li2019perspective,
  title={A perspective on off-policy evaluation in reinforcement learning},
  author={Li, Lihong},
  journal={Frontiers of Computer Science},
  volume={13},
  number={5},
  pages={911--912},
  year={2019},
  publisher={Springer}
}

@article{uehara2023future,
  title={Future-dependent value-based off-policy evaluation in pomdps},
  author={Uehara, Masatoshi and Kiyohara, Haruka and Bennett, Andrew and Chernozhukov, Victor and Jiang, Nan and Kallus, Nathan and Shi, Chengchun and Sun, Wen},
  journal={Advances in neural information processing systems},
  volume={36},
  pages={15991--16008},
  year={2023}
}

@article{mees2022calvin,
  title={Calvin: A benchmark for language-conditioned policy learning for long-horizon robot manipulation tasks},
  author={Mees, Oier and Hermann, Lukas and Rosete-Beas, Erick and Burgard, Wolfram},
  journal={IEEE Robotics and Automation Letters},
  volume={7},
  number={3},
  pages={7327--7334},
  year={2022},
  publisher={IEEE}
}

@article{zheng2025x,
    title={X-VLA: Soft-Prompted Transformer as Scalable Cross-Embodiment Vision-Language-Action Model},
    author={Zheng, Jinliang and Li, Jianxiong and Wang, Zhihao and Liu, Dongxiu and Kang, Xirui and Feng, Yuchun and Zheng, Yinan and Zou, Jiayin and Chen, Yilun and Zeng, Jia and others},
    journal={arXiv preprint arXiv:2510.10274},
    year={2025}
  }

@article{DBLP:journals/corr/abs-1906-01624,
  author       = {Alex Irpan and
                  Kanishka Rao and
                  Konstantinos Bousmalis and
                  Chris Harris and
                  Julian Ibarz and
                  Sergey Levine},
  title        = {Off-Policy Evaluation via Off-Policy Classification},
  journal      = {CoRR},
  volume       = {abs/1906.01624},
  year         = {2019},
  url          = {http://arxiv.org/abs/1906.01624},
  eprinttype   = {arXiv},
  eprint       = {1906.01624},
  bibsource    = {dblp computer science bibliography, https://dblp.org}
}

@InProceedings{black2025pi05,
  title     = {$\pi_{0.5}$: a Vision-Language-Action Model with Open-World Generalization},
  author    = {Black, Kevin and Brown, Noah and Darpinian, James and Dhabalia, Karan and Driess, Danny and Esmail, Adnan and Equi, Michael Robert and Finn, Chelsea and Fusai, Niccolo and Galliker, Manuel Y. and Ghosh, Dibya and Groom, Lachy and Hausman, Karol and ichter, brian and Jakubczak, Szymon and Jones, Tim and Ke, Liyiming and LeBlanc, Devin and Levine, Sergey and Li-Bell, Adrian and Mothukuri, Mohith and Nair, Suraj and Pertsch, Karl and Ren, Allen Z. and Shi, Lucy Xiaoyang and Smith, Laura and Springenberg, Jost Tobias and Stachowicz, Kyle and Tanner, James and Vuong, Quan and Walke, Homer and Walling, Anna and Wang, Haohuan and Yu, Lili and Zhilinsky, Ury},
  booktitle = {Proceedings of The 9th Conference on Robot Learning},
  pages     = {17--40},
  year      = {2025},
  editor    = {Lim, Joseph and Song, Shuran and Park, Hae-Won},
  volume    = {305},
  series    = {Proceedings of Machine Learning Research},
  publisher = {PMLR}
}

@article{jain2026weaver,
  title   = {{WEAVER}, Better, Faster, Longer: An Effective World Model for Robotic Manipulation},
  author  = {Jain, Arnav Kumar and Wu, Yilin and Farebrother, Jesse and Swamy, Gokul and Bajcsy, Andrea},
  journal = {arXiv preprint arXiv:2606.13672},
  year    = {2026}
}

@article{tseng2026sc3eval,
  title   = {{SC3-Eval}: Evaluating Robot Foundation Models via Self-Consistent Video Generation},
  author  = {Tseng, Wei-Cheng and Hussein, Gashon and Dong, Yuzhu and Ren, Allen Z. and Shi, Lucy X. and Wang, XuDong and Levine, Sergey and Li, Zhaoshuo and Gu, Jinwei and Shkurti, Florian and Liu, Ming-Yu and Vuong, Quan},
  journal = {arXiv preprint arXiv:2606.18610},
  year    = {2026}
}

@inproceedings{liang2026robometer,
  author    = {Anthony Liang and Yigit Korkmaz and Jiahui Zhang and Minyoung Hwang and Abrar Anwar and Sidhant Kaushik and Aditya Shah and Alex S. Huang and Luke Zettlemoyer and Dieter Fox and Yu Xiang and Anqi Li and Andreea Bobu and Abhishek Gupta and Stephen Tu and Erdem Biyik and Jesse Zhang},
  title     = {{Robometer: Scaling General-Purpose Robotic Reward Models via Trajectory Comparisons}},
  booktitle = {Proceedings of Robotics: Science and Systems},
  year      = {2026},
  address   = {Sydney, Australia},
  month     = {July},
  doi       = {10.15607/RSS.2026.XXII.140}
}

@inproceedings{guo2026ctrl,
  title={Ctrl-world: A controllable generative world model for robot manipulation},
  author={Guo, Yanjiang and Shi, Lucy and Chen, Jianyu and Finn, Chelsea},
  booktitle={International Conference on Learning Representations},
  volume={2026},
  pages={6121--6138},
  year={2026}
}

@article{ali2025world,
  title={World simulation with video foundation models for physical ai},
  author={Ali, Arslan and Bai, Junjie and Bala, Maciej and Balaji, Yogesh and Blakeman, Aaron and Cai, Tiffany and Cao, Jiaxin and Cao, Tianshi and Cha, Elizabeth and Chao, Yu-Wei and others},
  journal={arXiv preprint arXiv:2511.00062},
  year={2025}
}

@techreport{veoteam2025veo,
  author      = {{Veo Team}},
  title       = {{Veo}: a text-to-video generation system},
  institution = {Google DeepMind},
  year        = {2025},
  url         = {https://storage.googleapis.com/deepmind-media/veo/Veo-3-Tech-Report.pdf}
}

\clearpage
\appendix
\section{Appendix}
\label{sec:appendix}

This appendix provides four groups of supplementary experiments. Appendix~\ref{app:multimodal} evaluates RoboFQE on policies that condition on proprioceptive state in addition to images, a setting in which video world-model evaluators cannot operate. Appendix~\ref{app:realworld} reports results on real-world block-stacking tasks. Appendix~\ref{app:data_composition} examines how training data composition and encoder initialization jointly affect value estimation accuracy. Appendix~\ref{app:failure_modes} analyzes two failure modes of RoboFQE: sensitivity to dataset coverage and degradation under out-of-support policy actions.

\subsection{Multimodal Policy Inputs}
\label{app:multimodal}

The policies in the main experiments condition on visual observations and the language instruction. Because RoboFQE fits the action-value function on the policy input $x_t = f(h_t)$, it inherits the input modality of the target policy, and the FQE objective requires no change to accept proprioception. Pixel-space world-model evaluators instead generate image observations and cannot produce the proprioceptive channel that a proprioception-conditioned policy reads, so they cannot evaluate such a policy. This subsection evaluates RoboFQE on a policy that consumes proprioceptive state in addition to images across the four LIBERO suites.

The target policy extends the ResNet-T architecture from Section~\ref{sec:experimental_setup} with a proprioceptive input branch, receiving the 15-dimensional proprioceptive state provided by LIBERO (7 joint positions, end-effector position, end-effector orientation in axis-angle form, and gripper joint positions) together with the images and language instruction and outputting a 7-dimensional action. It is trained by behavioral cloning under the same pipeline as the vision-only policies. A single policy per suite is evaluated for value estimation, and three policies with separated success rates, taken at separate training stages, are compared for ranking. The Q-function follows Section~\ref{sec:robofqe_instantiation}: the proprioceptive state is projected through a learned embedding layer and concatenated with the DINOv3 visual features and the action embedding before language fusion. All FQE hyperparameters match the main LIBERO configuration, including $\gamma{=}0.99$ and the target-network synchronization period, so no separate search is performed for this setting.

Table~\ref{tab:multimodal} reports value estimation and ranking accuracy for the proprioception-conditioned policy. RoboFQE produces value estimates on all four suites, with absolute relative error ranging from 6.41\% on Spatial to 21.85\% on Long, and Pearson correlation against $V^\pi$ ranging from +0.80 to +0.93. The target policy here differs from the vision-only policies in Section~\ref{sec:experimental_setup}, so the magnitudes are not directly comparable to Table~\ref{tab:combined_results}. The two accuracy measures do not move together across suites: Long has the highest relative error but retains a correlation of +0.83, while Goal has a moderate relative error but the lowest rank correlation ($+0.80$), consistent with the separation between ranking and value calibration reported in Section~\ref{sec:main_results}. World-model evaluators are absent from Table~\ref{tab:multimodal} because image generation cannot produce the proprioceptive channel the policy reads.

\begin{table}[!ht]
\centering
\caption{Value estimation and policy ranking for a proprioception-conditioned ResNet-T policy on the four LIBERO suites, measured against the ground-truth discounted return $V^\pi$. All uncertainties are 95\% bootstrapped confidence intervals. Ground-truth per suite ($\gamma{=}0.99$): Spatial ($V^\pi{=}0.203$), Object ($V^\pi{=}0.154$), Goal ($V^\pi{=}0.256$), Long ($V^\pi{=}0.026$). World-model evaluators are omitted because image generation cannot produce the proprioceptive channel the policy reads.}
\label{tab:multimodal}
\vspace{0.5em}
\small
\setlength{\tabcolsep}{4pt}
\begin{tabular}{@{}lcccc@{}}
\toprule
Suite & ARE (\%) $\downarrow$ & MSE ($\times 10^{-5}$) $\downarrow$ & $r$ $\uparrow$ & MMRV $\downarrow$ \\
\midrule
Spatial & $6.41 \pm 0.87$  & $17.70 \pm 4.25$  & $+0.93$ & $0.02$ \\
Object  & $9.68 \pm 1.59$  & $23.60 \pm 7.85$  & $+0.90$ & $0.04$ \\
Goal    & $9.84 \pm 3.36$  & $82.83 \pm 49.74$ & $+0.80$ & $0.17$ \\
Long    & $21.85 \pm 5.82$ & $3.83 \pm 2.14$   & $+0.83$ & $0.17$ \\
\bottomrule
\end{tabular}
\end{table}

The coverage-dependent failure mode of FQE under out-of-support policy actions is analyzed in Appendix~\ref{app:failure_modes}, using the ranking set from this subsection.

\subsection{Real-World Setup Details}
\label{app:realworld}

This subsection details the real-world study reported in Section~\ref{sec:realworld}. The hardware is an SO-101 arm with two RGB cameras, a top camera and a wrist camera (Figure~\ref{fig:real_world_setup}); the policies condition on the two camera views and the language instruction without proprioceptive input. Three tabletop tasks are considered, each stacking one colored block on another: red on blue, blue on yellow, and yellow on red. The three tasks share the same red, yellow, and blue blocks and differ only in the source and target block. Each task is instantiated under five initial-position settings, and 10 teleoperated demonstrations are collected per setting, yielding 150 demonstrations in total with an average length of approximately 500 steps. To broaden the state-action coverage of the offline dataset, 100 additional trajectories are collected by rolling out a behavioral cloning checkpoint that is not among the target policies; their terminal success labels are annotated by LRM~\citep{wu2026large}.

The target policies are SmolVLA~\citep{shukor2025smolvla} and $\pi_{0.5}$~\citep{black2025pi05}, both fine-tuned on the demonstration data with LeRobot, together with a multitask behavioral cloning policy trained on the same data. Ground-truth returns are obtained by rolling out each policy on the robot for 100 episodes per task and recording task success manually, with the discount factor set to $\gamma{=}1$ so that the return of a policy equals its task success rate. The baselines are WorldGym~\citep{quevedo2026worldgym} and Cosmos 2.5-AC, both post-trained on the real-world offline dataset and applied as in the LIBERO experiments. RoboFQE is instantiated as in Section~\ref{sec:robofqe_instantiation}, using the DINOv3 encoder fine-tuned on the offline dataset with LRM-annotated terminal labels. Results are averaged over three seeds and reported as mean $\pm$ standard deviation in Table~\ref{tab:realworld_results}; with only three seeds, bootstrapped confidence intervals as in the LIBERO tables are not reported.

Unlike the LIBERO experiments, in which LRM is fine-tuned on the official demonstration data and achieves 97.4\% accuracy, the real-world setting applies LRM without task-specific fine-tuning. On the 250 real-world trajectories (150 demonstrations and 100 policy rollouts), zero-shot LRM achieves 83.3\% terminal annotation accuracy against manual labels. This lower accuracy increases the reward-inference error in the FQE training targets, making the real-world setting a stricter test of RoboFQE's robustness to imperfect reward annotation.

\subsection{Training Data Composition}
\label{app:data_composition}

The main experiments train a separate Q-function per LIBERO suite, using only the 750 trajectories collected for that suite. Because the four suites share a similar visual environment, trajectories from other suites provide additional state-action coverage in the same visual domain, although under different task instructions and success predicates. This subsection tests whether pooling off-task data from the remaining three suites into the FQE training set improves or degrades value estimation, and how this effect interacts with encoder initialization.

Two data conditions are compared. In the \textit{on-task} condition, FQE is trained on the 750 trajectories of the suite being evaluated, matching the main experiments. In the \textit{all-LIBERO} condition, FQE is trained on the combined dataset from all four suites (3\,000 trajectories, approximately 569k transitions). Each condition is run with both the DINOv3 encoder and the from-scratch ResNet-18 encoder, yielding a $2 \times 2$ comparison. The language instruction identifies each task, so the Q-function can discriminate across tasks through its language-conditioned architecture. All other hyperparameters match the main LIBERO configuration.

\begin{figure}[t]
\centering
\includegraphics[width=\textwidth]{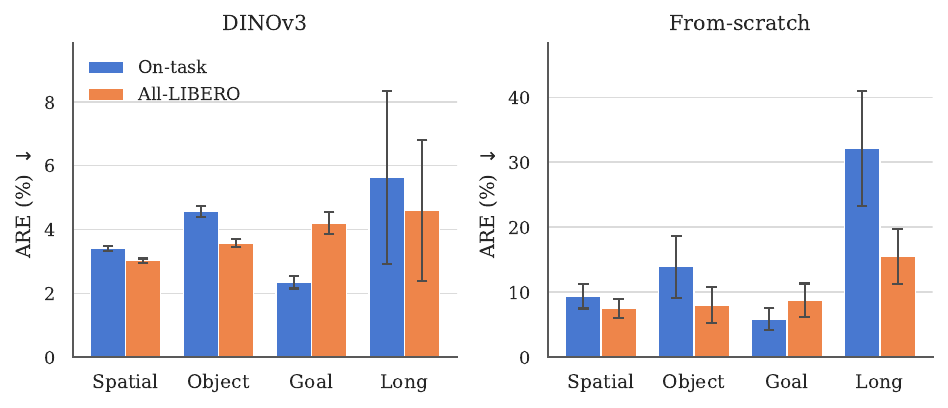}
\caption{Effect of training data composition and encoder initialization on value estimation accuracy across the four LIBERO suites. Bars report absolute relative error (ARE), with error bars giving 95\% bootstrapped confidence intervals. On-task trains FQE on the evaluated suite alone; all-LIBERO pools data from all four suites. Goal is the only suite in which pooling off-task data increases error under both encoders.}
\label{fig:data_composition}
\end{figure}

Figure~\ref{fig:data_composition} reports the results. With the DINOv3 encoder, pooling off-task data produces small ARE reductions on Spatial, Object, and Long (0.4--1.0 percentage points) but increases Goal's ARE from 2.34\% to 4.20\%. With the from-scratch encoder, the same pattern holds at larger scale: ARE on Long drops from 32.09\% to 15.49\% and Object from 13.89\% to 7.98\%, while Goal's ARE rises from 5.82\% to 8.74\%. The from-scratch encoder must learn visual representations entirely from the FQE training data, so broader visual coverage from off-task trajectories improves representation quality on the suites whose value estimation depends on discriminating between visually distinct scenes and objects (Spatial, Object) or sustaining accurate Bellman propagation over many steps (Long). Goal fixes the scene and objects across its 10 tasks and varies only the goal predicate specified by the language instruction; because its estimation difficulty lies in goal discrimination rather than visual diversity, additional visual coverage does not help, and the Q-function's fit to Goal-specific tasks degrades. This interaction reinforces the finding from Section~\ref{sec:main_results} that pretrained visual representations are the dominant factor for value estimation accuracy: once the encoder is sufficiently expressive, the marginal benefit of additional data diversity is small.

\subsection{Failure Modes of RoboFQE}
\label{app:failure_modes}

RoboFQE estimates a policy's value from a fixed offline dataset, so the dataset determines which policies the estimator can evaluate accurately. This subsection characterizes two conditions under which RoboFQE produces unreliable estimates. The first condition arises from the composition of the offline dataset; the second arises from the relationship between the evaluated policy and the data-collection policy. Both conditions are demonstrated on the LIBERO suites used in the main experiments.

\paragraph{Insufficient data coverage.}
Offline robot datasets are often built from teleoperated demonstrations, which are predominantly successful trajectories. When any data-driven evaluator is trained on such a dataset, it observes near-optimal behavior but few transitions that end in failure, limiting its ability to assess policies that fail. To measure this effect, RoboFQE is trained on the 50 human demonstrations per task alone and compared against the augmented dataset of the main experiments, which adds 25 rollouts per task from partially trained policies (Section~\ref{sec:experimental_setup}). Table~\ref{tab:data_augmentation} reports value estimation accuracy under the two conditions. Across all four suites, demonstration-only training increases ARE by 5--10$\times$ compared to the augmented dataset, with demonstration-only ARE highest on Long ($30.85\%$) and Object ($29.05\%$). Without failure transitions the Bellman targets at non-terminal steps are computed exclusively from high-value successor states, which biases the Q-function upward and prevents it from distinguishing between policies that succeed at different rates. Both world-model baselines also degrade under demonstration-only training, with ARE exceeding 100\% on multiple suites, consistent with the finding that video world models trained on success-skewed data cannot render plausible failure outcomes without rollout augmentation~\citep{tseng2025scalable}. Dataset augmentation with suboptimal rollouts reduces estimation error for all three evaluators, confirming that the coverage requirement is shared across evaluation paradigms rather than specific to value-based OPE.

\begin{table}[!ht]
\centering
\caption{Effect of dataset augmentation on RoboFQE value estimation across the four LIBERO suites. Demonstration-only uses the 50 human demonstrations per task; augmented adds 25 rollouts per task from partially trained policies. Both conditions use the DINOv3 encoder. Augmented results report 95\% bootstrapped confidence intervals; demonstration-only results are from a single training run.}
\label{tab:data_augmentation}
\vspace{0.5em}
\small
\setlength{\tabcolsep}{4pt}
\begin{tabular}{@{}llcc@{}}
\toprule
Data & Suite & ARE (\%) $\downarrow$ & MSE ($\times 10^{-5}$) $\downarrow$ \\
\midrule
\multirow{4}{*}{Demo-only}
 & Spatial & $19.22 \pm 2.61$  & $159.30 \pm 38.22$  \\
 & Object  & $29.05 \pm 4.78$  & $212.81 \pm 71.04$  \\
 & Goal    & $23.07 \pm 5.46$  & $298.47 \pm 130.37$ \\
 & Long    & $30.85 \pm 9.77$  & $8.05 \pm 4.75$     \\
\midrule
\multirow{4}{*}{Augmented}
 & Spatial & $3.40 \pm 0.08$   & $6.50 \pm 0.31$   \\
 & Object  & $4.56 \pm 0.17$   & $5.74 \pm 0.44$   \\
 & Goal    & $2.34 \pm 0.19$   & $3.38 \pm 0.59$   \\
 & Long    & $5.63 \pm 2.71$   & $0.42 \pm 0.38$   \\
\bottomrule
\end{tabular}
\end{table}

\paragraph{Out-of-support policy actions.}

\begin{figure}[!ht]
\centering
\includegraphics[width=0.65\textwidth]{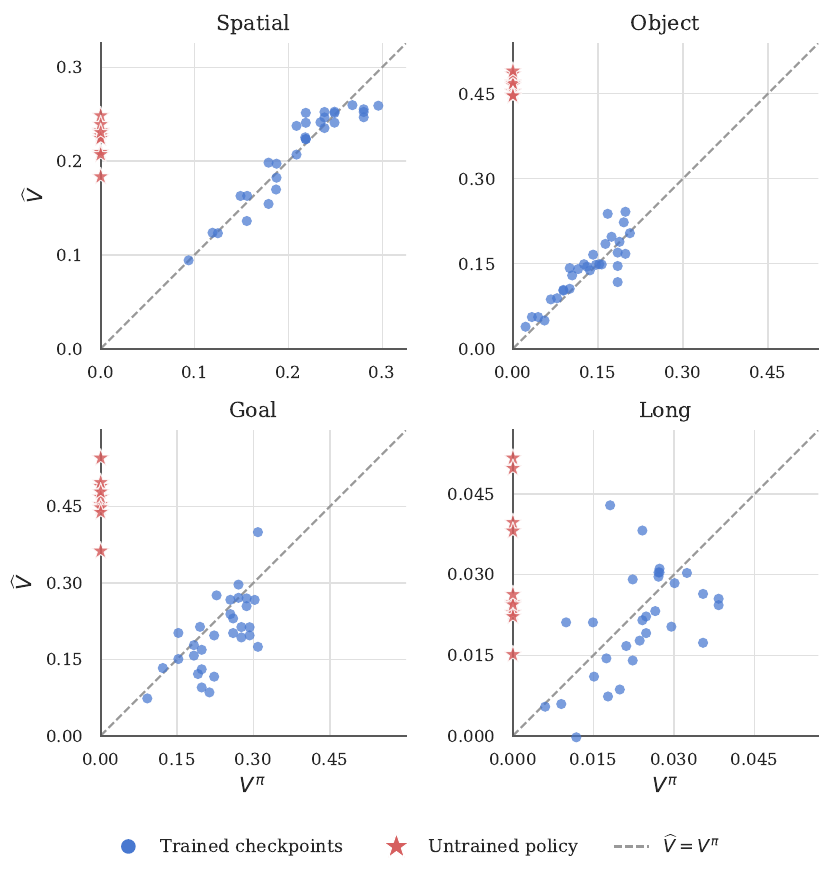}
\caption{Estimated versus ground-truth discounted return for each policy-task pair, one panel per LIBERO suite. Points show three trained checkpoints and one random policy, and the dashed line marks $\widehat{V}=V^\pi$.}
\label{fig:oos_scatter}
\end{figure}

RoboFQE evaluates a policy by querying the Q-function on the actions that policy selects. When those actions resemble the behavior recorded in the offline dataset, the Q-function interpolates within its training support and the estimate is reliable. When the policy selects actions the dataset does not cover, the Q-function extrapolates and can assign values with no basis in observed outcomes. To demonstrate this failure mode, the ranking set from Appendix~\ref{app:multimodal} is augmented with a random policy whose true success rate is zero, and the ranking metrics are recomputed. Table~\ref{tab:oos_ranking} reports the result. The random policy's actions are absent from the offline dataset, so the Q-function assigns it a positive value rather than a value near zero, placing it among the trained policies in estimated value and inverting pairwise orderings. The degradation is largest on Spatial: the Pearson correlation falls from $+0.93$ to $+0.54$ and the mean maximum rank violation rises from $0.02$ to $0.27$. This extrapolation error reflects a structural property of value-based OPE: the Q-function is trained under the offline data distribution, and its predictions carry no guarantee at state-action pairs at which that distribution assigns negligible density. This property bounds the intended use of RoboFQE: it produces reliable estimates for policies whose actions stay within the dataset's coverage, such as checkpoints from a shared training pipeline, and not for policies whose behavior departs sharply from the recorded data. This dependence on the overlap between the offline data and the target policy is a known property of off-policy evaluation, in which estimation error grows as the behavior and target policies diverge~\citep{voloshin2021empirical,fu2021benchmarks}.

\begin{table}[!ht]
\centering
\caption{Policy ranking degradation when a random policy with zero success rate is added to the ranked set, measured against the ground-truth discounted return $V^\pi$. The trained policies and all other settings match Table~\ref{tab:multimodal}.}
\label{tab:oos_ranking}
\vspace{0.5em}
\small
\setlength{\tabcolsep}{4pt}
\begin{tabular}{@{}lcccc@{}}
\toprule
 & \multicolumn{2}{c}{Without random} & \multicolumn{2}{c}{With random} \\
\cmidrule(lr){2-3} \cmidrule(lr){4-5}
Suite & $r$ $\uparrow$ & MMRV $\downarrow$ & $r$ $\uparrow$ & MMRV $\downarrow$ \\
\midrule
Spatial & $+0.93$ & $0.02$ & $+0.54$ & $0.27$ \\
Object  & $+0.90$ & $0.04$ & $+0.79$ & $0.15$ \\
Goal    & $+0.80$ & $0.17$ & $+0.75$ & $0.13$ \\
Long    & $+0.83$ & $0.17$ & $+0.70$ & $0.11$ \\
\bottomrule
\end{tabular}
\end{table}

Figure~\ref{fig:oos_scatter} shows this extrapolation directly, plotting the estimate against the ground-truth return for each policy-task pair in a representative ranking set, with one panel per suite. The trained checkpoints lie close to the diagonal across the full range of tasks, so their estimates remain accurate for policies whose actions are covered by the offline data. The random policy has a ground-truth return of zero on every task, yet the Q-function assigns it positive estimates that fall within or above the band of trained-checkpoint estimates: roughly $0.18$ to $0.25$ on Spatial, $0.36$ to $0.55$ on Object and Goal, and a similar pattern at the smaller value scale of Long. The spread across tasks is the central observation: the random policy's actions are absent from the offline data, so the Q-function extrapolates without constraint and assigns a different value on each task rather than one consistent estimate. This combination of overestimation and task-to-task variability ranks the random policy among the trained checkpoints inconsistently, producing the rank inversions reported in Table~\ref{tab:oos_ranking}.

\end{document}